%% file: main.tex
\documentclass[10pt,twocolumn,letterpaper]{article}

\usepackage[pagenumbers]{cvpr}
\input{preamble}
\definecolor{cvprblue}{rgb}{0.21,0.49,0.74}
\usepackage[pagebackref,breaklinks,colorlinks,allcolors=cvprblue]{hyperref}
\usepackage{pifont}
\usepackage{multirow}
\usepackage{xcolor}
\usepackage[dvipsnames,svgnames,x11names]{xcolor}
\usepackage{booktabs,multirow,graphicx,array,xcolor}

\usepackage[normalem]{ulem}      

\definecolor{BrightBlue}{HTML}{0000FF} 

\def\paperID{20047} 
\def\confName{CVPR}
\def\confYear{2026}

\title{MT-Mark: Rethinking Image Watermarking via Mutual-Teacher Collaboration with Adaptive Feature Modulation}

\author{
Fei Ge$^{1}$, Ying Huang$^{1}$, Jie Liu$^{1}$, Guixuan Zhang$^{1}$, Zhi Zeng$^{1}$, Shuwu Zhang$^{1}$, Hu Guan$^{2}$\\
{\small $^{1}$Beijing University of Posts and Telecommunications, $^{2}$Institute of Automation, Chinese Academy of Sciences}\\
{\tt\scriptsize \{fei.ge, ying.huang, AILJ, guixuan.zhang, zhi.zeng, shuwu.zhang\}@bupt.edu.cn, hu.guan@ia.ac.cn}
}

\begin{document}
\maketitle

\input{sec/0_abstract}

\input{sec/1_intro}

\input{sec/2_formatting}

\input{sec/3_finalcopy}

\input{sec/4_exa}

\input{sec/5_con}

{
    \small
    \bibliographystyle{ieeenat_fullname}

    \bibliography{main}
}


\end{document}

%% file: sec/0_abstract.tex
\begin{abstract}
Existing deep image watermarking methods follow a fixed embedding–distortion–extraction pipeline, where the embedder and extractor are weakly coupled through a final loss and optimized in isolation. This design lacks explicit collaboration, leaving no structured mechanism for the embedder to incorporate decoding-aware cues or for the extractor to guide embedding during training.
To address this architectural limitation, we rethink deep image watermarking by reformulating embedding and extraction as explicitly collaborative components. To realize this reformulation, we introduce a Collaborative Interaction Mechanism (CIM) that establishes direct, bidirectional communication between the embedder and extractor, enabling a mutual-teacher training paradigm and coordinated optimization.
Built upon this explicitly collaborative architecture, we further propose an Adaptive Feature Modulation Module (AFMM) to support effective interaction. AFMM enables content-aware feature regulation by decoupling modulation structure and strength, guiding watermark embedding toward stable image features while suppressing host interference during extraction. Under CIM, the AFMMs on both sides form a closed-loop collaboration that aligns embedding behavior with extraction objectives.
This architecture-level redesign changes how robustness is learned in watermarking systems. Rather than relying on exhaustive distortion simulation, robustness emerges from coordinated representation learning between embedding and extraction. Experiments on real-world and AI-generated datasets demonstrate that the proposed method consistently outperforms state-of-the-art approaches in watermark extraction accuracy while maintaining high perceptual quality, showing strong robustness and generalization.
\end{abstract}

%% file: sec/1_intro.tex
\section{Introduction}
\label{sec:intro}

\begin{figure}[t]
    \centering
    \includegraphics[width=\linewidth]{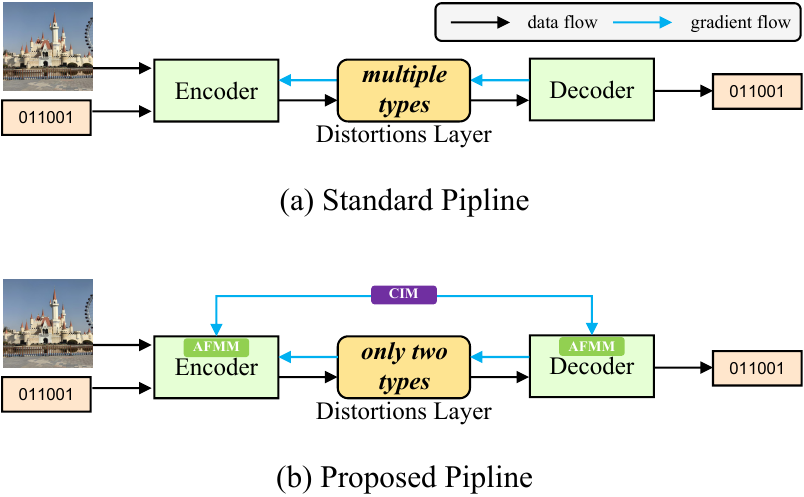}
    \caption{The structures of different image watermarking architectures.
(a) A standard embedding–distortion–extraction pipeline.
(b) MT-Mark, an explicitly collaborative architecture centered on the Collaborative Interaction Mechanism (CIM), which establishes mutual-teacher coordination between watermark embedding and extraction; Adaptive Feature Modulation Modules (AFMMs) provide the feature-level modulation interface for this collaboration.
Black arrows denote data flow, and blue arrows denote gradient flow.
}
    \label{fig:intro_concept}
\end{figure}

Digital image watermarking plays a critical role in copyright protection and content authentication, where robustness against diverse and unpredictable distortions is a fundamental requirement~\cite{kušen2017analysistwitterdiscussion2016}. In practical scenarios, watermarked images are frequently subjected to compression, noise, geometric transformations, and even compound or unforeseen perturbations during storage, transmission, and redistribution. Accurately recovering embedded watermarks under such conditions remains a long-standing challenge.

As illustrated in \cref{fig:intro_concept}(a), most deep-learning-based watermarking methods adopt a fixed end-to-end pipeline consisting of three sequential stages: embedding, distortion simulation, and extraction~\cite{zhu2018hidden,jia2021mbrs,fang2022pimog}. To improve robustness, existing approaches typically rely on extensive data augmentation, incorporating a wide range of simulated distortions during training~\cite{Fernandez_2023_ICCV,ci2024wmadapteraddingwatermarkcontrol,guo2024freqmark,zhang2024trainingfreeplugandplaywatermarkframework,sander2025watermarklocalizedmessages,fernandez2024videosealopenefficient,zhang2023editguardversatileimagewatermarking,zhang2025omniguardhybridmanipulationlocalization,lu2025robustwatermarkingusinggenerative,sauer2023adversarialdiffusiondistillation}. While effective to some extent, this paradigm suffers from two fundamental limitations.

First, the distortion models used during training can only approximate a limited subset of real-world degradations. As a result, models often overfit specific distortion patterns rather than learning genuinely distortion-invariant representations, leading to a significant performance drop under unseen perturbations. Second, and more critically, existing watermarking frameworks exhibit an inherent architectural limitation: the embedder and extractor are weakly coupled through a final scalar loss and optimized largely in isolation. This design lacks an explicit mechanism for collaboration, leaving no structured way for the embedder to incorporate decoding-aware cues or for the extractor to guide embedding behavior during training. Consequently, the interaction between embedding and extraction is confined to loss-level supervision, which fundamentally limits coordinated optimization and constrains robustness learning.

Motivated by this architectural bottleneck, we rethink deep image watermarking from an architectural perspective. Instead of treating embedding and extraction as isolated sequential stages, we reformulate them as explicitly collaborative components within a single framework. To realize this reformulation, we propose a \textbf{Collaborative Interaction Mechanism (CIM)} that establishes direct, bidirectional communication between the embedder and extractor. Through CIM, the two components are trained in a \emph{mutual-teacher} paradigm, where embedding and extraction continuously exchange informative signals and jointly optimize their behaviors.

Built upon this explicitly collaborative architecture, we further introduce an \textbf{Adaptive Feature Modulation Module (AFMM)} to support effective interaction. AFMM enables content-aware feature regulation by decoupling modulation structure and modulation strength. On the embedding side, AFMM guides watermark injection toward structurally stable and distortion-resilient image features; on the extraction side, it suppresses host interference and enhances watermark-related cues. Under the coordination of CIM, the AFMMs in the embedder and extractor form a closed-loop collaboration, aligning embedding strategies with extraction objectives at the feature level.

This architecture-level redesign fundamentally changes how robustness is learned in deep watermarking systems. Rather than relying on exhaustive distortion simulation, robustness emerges from coordinated representation learning between embedding and extraction. As shown in \cref{fig:intro_concept}(b), this collaborative architecture enables effective robustness learning even when trained with only two representative distortions. Extensive experiments on real-world and AI-generated image datasets demonstrate that the proposed framework, termed \textbf{MT-Mark}, consistently outperforms state-of-the-art methods in watermark extraction accuracy while maintaining high perceptual fidelity.

Our main contributions are summarized as follows:
\begin{itemize}
    \item[\ding{111}] We rethink the architecture of deep image watermarking and propose an explicitly collaborative framework that tightly couples embedding and extraction through direct interaction.
    \item[\ding{111}] We design a Collaborative Interaction Mechanism (CIM) that establishes bidirectional information flow between the embedder and extractor, enabling a mutual-teacher training paradigm.
    \item[\ding{111}] We introduce an Adaptive Feature Modulation Module (AFMM) to support collaboration by enabling content-aware and distortion-resilient feature regulation.
    \item[\ding{111}] Extensive experiments show that MT-Mark achieves an average watermark extraction accuracy improvement of at least 4.47\% over state-of-the-art methods across diverse distortion scenarios, while maintaining excellent perceptual quality.
\end{itemize}

\begin{figure*}[t]
\centering
\includegraphics[width=\textwidth]{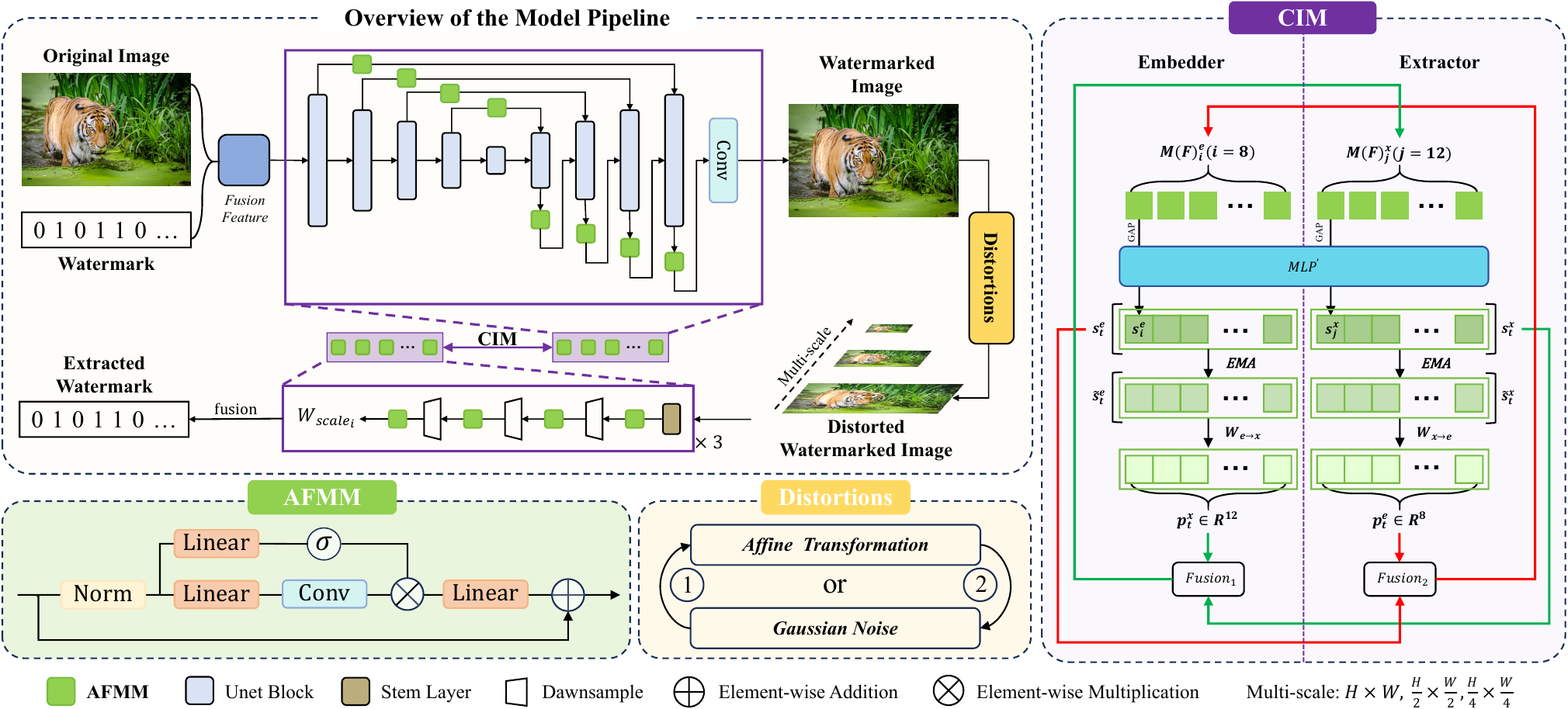} 
\caption{Overall framework of the proposed watermarking method MT-Mark. The top-left illustrates the overall pipeline, including watermark embedder, distortions, and extractor. The bottom-left presents detailed structures of key components, including the AFMM module and distortions layer. The right part shows the detailed structure of the CIM.}
\label{框架图}
\end{figure*}

%% file: sec/2_formatting.tex
\section{Related Work}
\label{sec:related}

Robust image watermarking was originally developed for copyright protection and image provenance tracking~\cite{van1994digital}. Early methods relied on hand-crafted designs in either the spatial domain, where watermark signals are directly embedded into pixel values~\cite{nikolaidis1998robust}, or the frequency domain, where transform coefficients are modulated using Fourier, wavelet, or cosine transforms~\cite{rahman2013dwt,10045698,8673925}. With the rise of deep learning, watermarking research has largely shifted toward end-to-end learning-based frameworks, which significantly improve imperceptibility and robustness through data-driven representations.

Most existing deep watermarking methods adopt a fixed end-to-end architecture following an \emph{embedding--distortion--extraction} pipeline~\cite{fernandez2022watermarkingimagesselfsupervisedlatent,bui2023trustmarkuniversalwatermarkingarbitrary,hu2025maskimagewatermarking,hu2024robustwiderobustwatermarkinginstructiondriven}. These methods are typically trained with a composite objective that balances image fidelity (e.g., MSE, LPIPS) and watermark decoding accuracy (e.g., BCE)~\cite{meng2024latentwatermarkinjectdetect,sander2025watermarklocalizedmessages,jovanović2025watermarkingautoregressiveimagegeneration}. To enhance robustness, a prevalent strategy is distortion simulation, where a wide range of image degradations are introduced during training~\cite{zhu2018hidden,fernandez2022watermarkingimagesselfsupervisedlatent,bui2023rosteals,rezaei2024lawa}. While effective against seen perturbations, this paradigm relies heavily on the coverage of the distortion set and often struggles to generalize to unseen distortions.

More importantly, these approaches share a common architectural characteristic: the embedder and extractor are weakly coupled through a final scalar loss and optimized in isolation. This loss-only coupling provides no explicit mechanism for fine-grained interaction between embedding and extraction during training, leaving embedding strategies unaware of decoding behavior and limiting coordinated representation learning.

%% file: sec/3_finalcopy.tex
\section{Methodology}

\subsection{Overview}

The overall framework of \textbf{MT-Mark} is illustrated in \cref{框架图}. 
MT-Mark consists of a watermark embedder, a watermark extractor, a distortions layer used during training, and a \textbf{Collaborative Interaction Mechanism (CIM)} that explicitly couples the embedder and extractor.

Given an original image $I$ and a binary watermark $w$, the embedder injects $w$ into $I$ to generate a watermarked image $I_{\mathrm{wm}}$. 
The watermarked image is then passed through a distortions layer, which applies representative image degradations during training. 
The extractor takes the distorted image as input and produces an estimated watermark $\hat{w}$. 
The embedder and extractor are connected by CIM, which enables bidirectional information exchange during training.

To support feature-level interaction under CIM, both the embedder and extractor are equipped with \textbf{Adaptive Feature Modulation Modules (AFMMs)}. 
AFMMs provide a structured mechanism for regulating intermediate feature representations and serve as the carriers through which CIM exchanges and adapts modulation states between the two components. 
Through this design, the embedder and extractor can coordinate their behaviors via feature-level feedback while maintaining their respective functional roles.

The entire framework is trained end-to-end using a composite loss function:
\begin{equation}
\mathcal{L} = \lambda_{1}\operatorname{BCE}\bigl(\hat{w}, w\bigr)
+ \lambda_{2}\operatorname{MSE}\bigl(I_{\mathrm{wm}}, I\bigr)
+ \lambda_{3}\operatorname{LPIPS}\bigl(I_{\mathrm{wm}}, I\bigr),
\end{equation}
where the BCE term encourages accurate watermark recovery, while the MSE and LPIPS terms jointly preserve the visual fidelity of the watermarked image. 
The weighting coefficients $\lambda_{1}$, $\lambda_{2}$, and $\lambda_{3}$ control the balance between robustness and imperceptibility.

\subsection{Collaborative Interaction Mechanism}

CIM enables bidirectional, feature-level interaction between the watermark embedder and extractor during training.
By exchanging and adapting intermediate modulation information, the two components act as mutual teachers, using feedback from each other to better perform their respective tasks.

CIM interfaces with the \textbf{Adaptive Feature Modulation Modules (AFMMs)} embedded in both the embedder and extractor.
AFMMs serve as structured carriers for interaction, as their modulation states explicitly encode how feature representations are being adjusted during watermark embedding and extraction.
By exchanging and adapting these modulation states, CIM aligns embedding strategies with extraction behavior in a controlled and interpretable manner.

\noindent\textbf{Shared preliminary steps: state extraction, smoothing, and cross-end mapping.}
Each AFMM produces a modulation map \(M(F)^e_i\) on the embedder side (\(i=1,\dots,8\)) and \(M(F)^x_j\) on the extractor side (\(j=1,\dots,12\)), which reflects its current feature adjustment behavior.
These modulation maps are spatially compressed via global average pooling and passed through a weight-shared MLP, denoted as \( \mathrm{MLP}' \), to obtain instantaneous scalar states:
\begin{equation}
s^e_i = \mathrm{MLP}'\!\left(\mathrm{GAP}(M(F)^e_i)\right), \quad
s^x_j = \mathrm{MLP}'\!\left(\mathrm{GAP}(M(F)^x_j)\right).
\end{equation}

The instantaneous state vectors \(s^e_t \in \mathbb{R}^{8}\) and \(s^x_t \in \mathbb{R}^{12}\) are then smoothed using an exponential moving average to capture stable historical trends:
\begin{equation}
\tilde{s}^e_t = \beta\,\tilde{s}^e_{t-1} + (1-\beta)\,s^e_t, \quad
\tilde{s}^x_t = \beta\,\tilde{s}^x_{t-1} + (1-\beta)\,s^x_t,
\end{equation}
where \(\beta\) controls the balance between stability and responsiveness.
The smoothed states are linearly projected to generate cross-end guidance signals:
\begin{equation}
p^x_t = W_{e\to x}\,\tilde{s}^e_t, \quad
p^e_t = W_{x\to e}\,\tilde{s}^x_t,
\end{equation}
where \(W_{e\to x}\) and \(W_{x\to e}\) are learnable projection matrices.
Here, \(p^e_{t,i}\) and \(p^x_{t,j}\) denote the guidance signals provided to the \(i\)-th embedder AFMM and the \(j\)-th extractor AFMM, respectively.

\noindent\textbf{Fusion and modulation (embedder side; extractor symmetric).}
For the \(i\)-th AFMM in the embedder, CIM combines its internal state \(s^e_i\) with the external guidance \(p^e_{t,i}\) from the extractor using a learnable gating mechanism:
\begin{equation}
\alpha^e_i = \sigma(\lambda^e_i), \quad
g^e_i = \alpha^e_i \cdot p^e_{t,i} + (1-\alpha^e_i)\cdot s^e_i,
\end{equation}
where \(\sigma(\cdot)\) is the sigmoid function and \(\lambda^e_i\) is a learnable parameter controlling the relative influence of cross-end feedback and self-state.

The fused scalar \(g^e_i\) is then expanded into a channel-wise adjustment vector:
\begin{equation}
a^e_i = w^e_i \cdot g^e_i + b^e_i,
\end{equation}
where \(w^e_i, b^e_i \in \mathbb{R}^{C}\) are learnable per-channel scale and bias parameters.
The adjustment vector is broadcast spatially and added to the original modulation map, yielding the corrected feature update:
\begin{equation}
F^{e}_{\mathrm{out},i} = F^{e}_i + \bigl(M(F)^e_i + a^e_i\bigr).
\end{equation}

This incremental, channel-wise modulation preserves spatial consistency while selectively amplifying or attenuating feature adjustments based on both local modulation intent and cross-end feedback.
The extractor side follows the same procedure symmetrically, using \(p^x_{t,j}\), \(s^x_j\), and its own gating and channel-wise parameters to update \(F^{x}_{\mathrm{out},j}\).

By enabling co-adaptive, feature-level coordination between the embedder and extractor, CIM supports robust and reliable watermark embedding and extraction.
Next, we describe AFMM, which produces the modulation states used by CIM and provides the feature-level interface for interaction.

\subsection{Adaptive Feature Modulation Module}

In image watermarking, distortions affect different regions unequally. Inspired by the gating mechanism in large language models~\cite{dauphin2017languagemodelinggatedconvolutional}, we employ AFMM to adaptively refine feature maps via separate \emph{strength and shape} branches.

Given an input feature $F$, we first normalize it to stabilize its statistics:
\begin{equation}
\hat{F} = \operatorname{Norm}(F)
\end{equation}
This stabilizes feature statistics and facilitates modulation learning.

The normalized feature is then processed by two parallel paths. \emph{The strength branch} produces a bounded coefficient that controls the magnitude and selectivity of the adjustment:
\begin{equation}
c = \sigma(W_1 \hat{F} + b_1),
\end{equation}
where \(\sigma\) is a bounded nonlinearity (e.g., sigmoid) to prevent excessive perturbation in sensitive regions and allow stronger adjustment where appropriate. \emph{The shape branch} generates a localized deformation that carries the target modification:
\begin{equation}
t = \operatorname{Conv}(W_2 \hat{F} + b_2),
\end{equation}
capturing how the feature should be altered in structure to embed the desired signal in a form resilient to distortion.

The two outputs interact element-wise and are fused via a projection to form the refinement residual:
\begin{equation}
M(F) = W_o (c \odot t) + b_o,
\end{equation}
where the multiplicative coupling enables content-adaptive modulation: the shape transformation \(t\) is selectively emphasized or attenuated spatially and across channels by \(c\). Finally, the residual is added back to the original feature:
\begin{equation}
F_{\mathrm{out}} = F + M(F).
\end{equation}

This design explicitly separates \emph{what form} the modification takes (via $t$) from \emph{how strongly and where} it is applied (via $c$), achieving adaptive, localized refinement while preserving the base representation through the residual connection.

\subsection{Watermark Embedder}

The embedding pipeline first constructs a fused feature \(F_{\mathrm{fuse}}\) from three distinct streams. To provide initial resilience to geometric distortions, parallel Deformable Convolutions (DCNs) extract global and local image feature; concurrently, the watermark \(w\) is reshaped into a feature map. These streams are losslessly concatenated to form a joint representation for subsequent adaptive processing.

The fused feature is fed into a U-Net based encoder–decoder backbone, where watermark injection is realized via a two-stage AFMMs along skip and reconstruction paths. Specifically, for the high-resolution feature \(x_i\) from the \(i\)-th encoder layer (before downsampling), an AFMM modulates it to produce the skip connection feature:
\begin{equation}
x_i' = x_i + M_i(x_i),
\end{equation}
where \(M_i(\cdot)\) denotes the AFMM refinement term at that encoder stage, introducing content-adaptive perturbations on rich-detail branches.

On the decoder side, let \(d_{i+1}\) be the output from the previous scale; it is first upsampled:
\begin{equation}
d_{i+1}^{\uparrow} = \operatorname{Upsample}(d_{i+1}).
\end{equation}
The upsampled feature is then fused with the modulated skip feature \(x_i'\) via channel-wise concatenation followed by convolution, as in standard U-Net skip connection:
\begin{equation}
y_i = \operatorname{Conv}\big([\;d_{i+1}^{\uparrow};\; x_i'\;]\big).
\end{equation}
The fused representation is further refined by a decoder-side AFMM:
\begin{equation}
y_i' = y_i + M'_i(y_i),
\end{equation}
where \(M'_i(\cdot)\) is the corresponding AFMM refinement term, reinforcing and harmonizing the injected watermark perturbations with the reconstruction feature.

This “inject on skip branch, then reinforce during reconstruction” two-stage embedding creates a redundant, content-aware watermark distribution across scales while preserving the image structure. After the final upsampling, the refined feature \(d_{\mathrm{final}}'\) is projected back to image space by a terminal \(3\times 3\) convolution:
\begin{equation}
I_{\mathrm{wm}} = \psi(d_{\mathrm{final}}'),
\end{equation}
where \(\psi(\cdot)\) denotes the \(3\times 3\) conv, optionally followed by a bounded activation function tanh, that maps to RGB, yielding the watermarked image \(I_{\mathrm{wm}} \in \mathbb{R}^{H\times W\times 3}\).

\subsection{Distortion Layer}

To enhance robustness, our model is trained with two representative distortions: one geometric and one signal processing based. We chose the Affine transform as it serves as a comprehensive proxy for the entire class of geometric attacks, while Gaussian noise represents a fundamental, unstructured corruption in the signal processing domain. For the geometric class, we adopt the \textbf{Affine transform}, covering rotation, translation, scaling, and shearing. Sampling from continuous ranges promotes generalization beyond fixed transformations. For signal processing distortion, we use \textbf{Gaussian noise}, a fundamental and irregular form of corruption. Resisting this type of distortion encourages the model to develop robust feature representations that generalize to other degradations such as blur or compression.

During training, both distortions are applied in a random order to avoid order-specific bias. Affine parameters are uniformly sampled as follows: rotation ($[-90^{\circ}, 90^{\circ}]$), translation ($[-0.3, 0.3]$), scaling ($[0.8, 1.2]$), and shearing ($[-30^{\circ}, 30^{\circ}]$). Gaussian noise is applied per pixel with $\mu=0$ and $\sigma=0.04$. The application order is randomized by $\pi \sim \mathrm{Bernoulli}(0.5)$.

\subsection{Watermark Extractor}

The extractor uses multi-scale inputs (full, 1/2, 1/4) and cascaded AFMM refinement to recover watermarks under complex distortions, with cross-scale redundancy compensating for local degradations.

At each scale \(s\), the input \(I_s\) is first processed by a stem layer to obtain an initial feature, and then fed into a cascade of four AFMM refinement stages: the first three stages are each followed by a downsampling, and the final stage performs a last AFMM-based refinement at the coarsest resolution. This design ensures that at each resolution an AFMM explicitly suppresses host interference and strengthens the residual watermark signal, while the subsequent downsampling progressively expands the receptive field and aggregates broader context, yielding a purified and robust representation. Denoting this cascade by \(\mathcal{F}_s\), the output at scale \(s\) is
\begin{equation}
u_s = \mathcal{F}_s(I_s).
\end{equation}

The three scale-specific outputs are concatenated and passed through a multi-layer perceptron (MLP) for fusion, followed by a linear projection with sigmoid activation to produce the estimated watermark probability vector:
\begin{equation}
\hat{w} = \sigma\!\left(W_{\mathrm{out}}\;\mathrm{MLP}([u_1;\; u_{1/2};\; u_{1/4}]) + b_{\mathrm{out}}\right),
\end{equation}
where \(\hat{w}\in(0,1)^l\) gives per-bit prediction probabilities. By fusing multi-scale features nonlinearly, the MLP mitigates information loss when individual scales degrade. The extractor combines cascaded AFMM refinement and scale redundancy to progressively suppress interference and enhance extraction robustness under complex distortions.

%% file: sec/4_exa.tex



\section{Experiment and Evaluation}

\subsection{Implementation Details and Metrics}

We train our model on the COCO dataset, randomly sampling 100{,}000 images. All input images are resized to a fixed resolution \(h \times w = 256 \times 256\), and each image is embedded with a randomly generated watermark of length 128 bits. The loss weights are set to \(\lambda_{1}=1.5\), \(\lambda_{2}=1.0\), and \(\lambda_{3}=1.0\). Training is performed on a single NVIDIA RTX 4090 GPU with a batch size of 16 for 75 epochs. 

For image quality assessment, we report the widely used Peak Signal-to-Noise Ratio (PSNR), Structural Similarity Index (SSIM), and Learned Perceptual Image Patch Similarity (LPIPS), and additionally introduce the Color-Vision Visual Difference Predictor (CVVDP)~\cite{Mantiuk_2024}, which models human sensitivity to color differences via a perceptual detection threshold, better capturing subjective color distortion after watermark embedding. To evaluate robustness and generalization, we measure the bit accuracy of the extracted watermark (Acc), defined as the proportion of correctly recovered bits under various distortions.

We benchmark our approach against several SOTA methods: DCTDWT~\cite{al2007combined}, HiDDeN~\cite{zhu2018hidden}, SSL~\cite{fernandez2022watermarkingimagesselfsupervisedlatent}, MBRS~\cite{jia2021mbrs}, 
RoSteALS~\cite{bui2023rosteals},
TrustMark~\cite{bui2023trustmarkuniversalwatermarkingarbitrary}, and LaWa~\cite{rezaei2024lawa}, covering frequency‐domain, end‐to‐end deep network, and generative‐model watermarking paradigms. For all competitors, we use their official open-source implementations and released pretrained weights, following the authors' recommended evaluation settings whenever available. We comprehensively evaluate all methods on three datasets, COCO (real low-resolution)~\cite{lin2014microsoft}, DIV2K (real high-resolution)~\cite{agustsson2017ntire} and Chameleon (AI-generated)~\cite{yan2025sanitycheckaigeneratedimage}, in terms of image quality, watermark robustness, and generalization.

In addition, we conduct ablation studies to assess the contributions of key components, including AFMM, CIM, the distortions layer, and the extractor’s multi-scale design.

\begin{table}[t]
  \centering
  \small
  \caption{Evaluation of the watermark imperceptibility. We report PSNR, SSIM, LPIPS and CVVDP between watermarked and original images of COCO, DIV2K and Chameleon. The best results are highlighted in {\color{Red}\textbf{red}}, and the second one is {\color{BrightBlue}\underline{blue}}.}
  \label{tab:reconstruction_quality}
  \setlength{\tabcolsep}{4.9pt}
  \begin{tabular}{>{\centering\arraybackslash}m{0.01mm} l c c c c}
    \toprule[1.2pt]
    & Method 
    & PSNR $\uparrow$ 
    & SSIM $\uparrow$ 
    & LPIPS $\downarrow$ 
    & CVVDP $\uparrow$ \\
    \midrule
    \multirow{8}{*}{\rotatebox{90}{\textbf{COCO}}}
      & DCTDWT~\cite{rahman2013dwt}     & 17.2518 & 0.8791  & 0.1977 & 7.0042 \\
      & HiDDeN~\cite{zhu2018hidden}     & 28.5372 & 0.8518  & 0.0949 & 9.1082 \\
      & MBRS~\cite{jia2021mbrs}         & {\color{BrightBlue}\underline{39.2089}} & {\color{Red}\textbf{0.9820}} & 0.0180 & {\color{BrightBlue}\underline{9.7003}} \\
      & SSL~\cite{fernandez2022watermarkingimagesselfsupervisedlatent} & 37.8068 & 0.9646  & 0.0703 & 8.8098 \\
      & RoSteALS~\cite{bui2023rosteals} & 28.1416 & 0.8414  & 0.0422 & 8.3549 \\
      & Trustmark~\cite{bui2023trustmarkuniversalwatermarkingarbitrary} 
                                          & {\color{Red}\textbf{39.9301}} & 0.9718 & {\color{BrightBlue}\underline{0.0138}} & 9.5890 \\
      & LaWa~\cite{rezaei2024lawa}      & 31.6856 & 0.6997  & 0.0756 & 7.6515 \\
      & $\text{MT-Mark}_{\text{(ours)}}$                            & 38.1006 & {\color{BrightBlue}\underline{0.9726}} & {\color{Red}\textbf{0.0085}} & {\color{Red}\textbf{9.8781}} \\
    \midrule
    \multirow{8}{*}{\rotatebox{90}{\textbf{DIV2K}}}
      & DCTDWT~\cite{rahman2013dwt}     & 28.7996 & 0.9376  & 0.0241 & 9.2123 \\
      & HiDDeN~\cite{zhu2018hidden}     & 30.2489 & 0.9511  & 0.0653 & 8.5779 \\
      & MBRS~\cite{jia2021mbrs}         & {\color{BrightBlue}\underline{38.7218}} & {\color{Red}\textbf{0.9907}} & 0.0146 & {\color{BrightBlue}\underline{9.7612}} \\
      & SSL~\cite{fernandez2022watermarkingimagesselfsupervisedlatent} & 36.3833 & 0.9680  & 0.0933 & 8.9579 \\
      & RoSteALS~\cite{bui2023rosteals} & 26.4842 & 0.8355  & 0.0501 & 6.3214 \\
      & Trustmark~\cite{bui2023trustmarkuniversalwatermarkingarbitrary} 
                                          & 38.7000 & 0.9711 & {\color{BrightBlue}\underline{0.0136}} & 9.5614 \\
      & LaWa~\cite{rezaei2024lawa}      & 30.5676 & 0.6536  & 0.0927 & 5.6031 \\
      & $\text{MT-Mark}_{\text{(ours)}}$                            & {\color{Red}\textbf{38.7223}} & {\color{BrightBlue}\underline{0.9746}} & {\color{Red}\textbf{0.0135}} & {\color{Red}\textbf{9.8108}} \\
    \midrule
    \multirow{8}{*}{\rotatebox{90}{\textbf{Chameleon}}}
      & DCTDWT~\cite{rahman2013dwt}     & 28.1619 & 0.9360 & 0.1083 & 7.6198 \\
      & HiDDeN~\cite{zhu2018hidden}     & 31.7328 & 0.9420 & 0.0677 & 8.9922 \\
      & MBRS~\cite{jia2021mbrs}         & 38.5991 & {\color{BrightBlue}\underline{0.9733}} & 0.0220 & {\color{BrightBlue}\underline{9.6058}} \\
      & SSL~\cite{fernandez2022watermarkingimagesselfsupervisedlatent} & 36.1513 & 0.9688 & 0.1056 & 8.8128 \\
      & RoSteALS~\cite{bui2023rosteals} & 31.0424 & 0.9326 & 0.0272 & 8.2453 \\
      & Trustmark~\cite{bui2023trustmarkuniversalwatermarkingarbitrary} 
                                          & {\color{BrightBlue}\underline{39.1901}} & 0.9627 & {\color{BrightBlue}\underline{0.0132}} & 9.5681 \\
      & LaWa~\cite{rezaei2024lawa}      & 31.9834 & 0.8147 & 0.0569 & 7.4639 \\
      & $\text{MT-Mark}_{\text{(ours)}}$                            & {\color{Red}\textbf{40.0436}} & {\color{Red}\textbf{0.9735}} & {\color{Red}\textbf{0.0123}} & {\color{Red}\textbf{9.8661}} \\
    \bottomrule[1.2pt]
  \end{tabular}
\end{table}

\begin{figure*}[t]
  \centering
  \includegraphics[width=\linewidth]{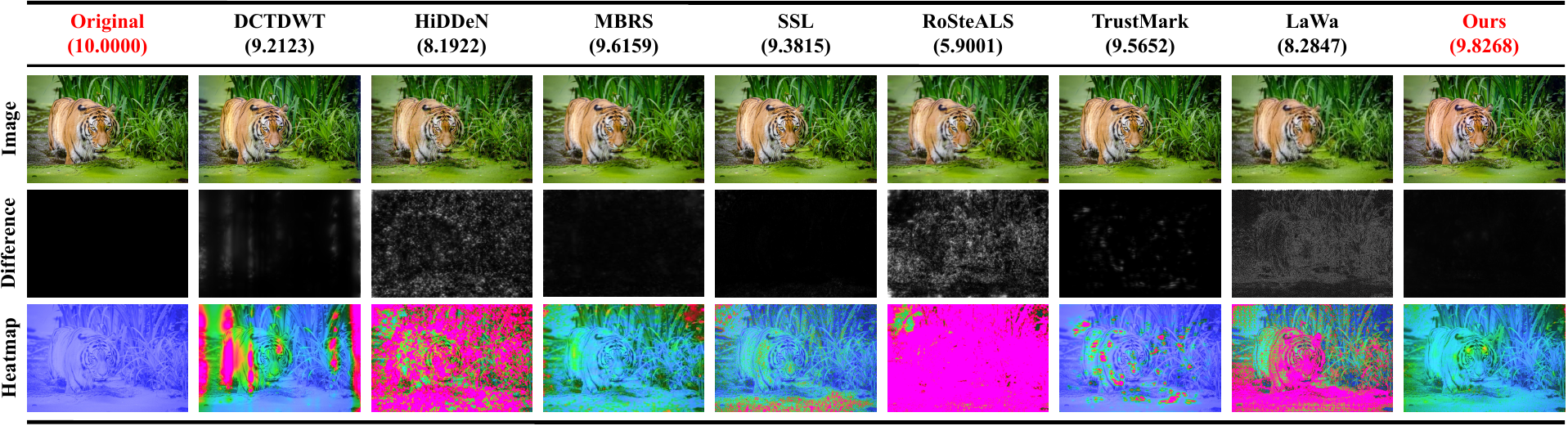} 
  \caption{Visual comparison on a DIV2K sample. \textbf{Top:} original and watermarked images (CVVDP score above each). \textbf{Middle:} difference maps. \textbf{Bottom:} CVVDP heatmaps (blue/green = imperceptible, pink = visible).}
  \label{cvvdp_heatmap}
\end{figure*}

\subsection{Perceptual Quality}
\label{sec:quality}

\cref{tab:reconstruction_quality} reports the perceptual quality of watermarked images. While traditional metrics like PSNR, SSIM, and LPIPS are provided, they do not fully capture the subtle color shifts that are critical to human-perceived fidelity. Standard LPIPS, for instance, is excellent at measuring texture and structure loss but is less specialized in modeling human sensitivity to color deviations. To address this, we introduce CVVDP, a metric specifically designed to model the human visual system’s varying sensitivity to color differences, making it a more stringent and accurate measure of high-fidelity color imperceptibility.

On COCO, while MBRS and TrustMark achieve high SSIM and PSNR scores, our method attains the best perceptual fidelity, with the lowest LPIPS (0.0085) and highest CVVDP (9.8781). The trend holds on the high-resolution DIV2K and AI-generated Chameleon datasets, where our method surpasses all baselines across all four metrics. This performance validates that, under the coordination of CIM, AFMMs enable perceptually aligned watermark embedding across diverse image sources and resolutions.

\cref{cvvdp_heatmap} provides a crucial visual comparison: the difference maps (residuals) reveal the limits of pixel-wise comparisons—while methods like LaWa exhibit obvious artifacts, the maps for top methods such as MBRS, TrustMark, and Ours are nearly black and visually indistinguishable—therefore the `Bottom: CVVDP heatmaps` are essential, translating subtle color shifts that are invisible in residuals (and not fully differentiated by other metrics) into a clear visual representation. In these heatmaps, blue indicates imperceptible differences, while green-to-pink signifies increasingly visible color errors. The heatmap for our method (Ours, 9.8268) is almost entirely cool blue, proving negligible perceptual impact. In contrast, other high-scoring methods like MBRS (9.6159), TrustMark (9.5652), and SSL (9.3815) still reveal localized green/cyan spots on the tiger’s fur and grass edges, indicating small but detectable color deviations. Methods like HiDDeN (8.1922) and RoSteALS (5.9001) show large patches of bright pink, confirming the severe color distortion reported in \cref{tab:reconstruction_quality}. This qualitative–quantitative alignment confirms our method’s superior perceptual imperceptibility.

\begin{table*}
  \centering
  \small
  \caption{Bit accuracy under geometric distortions (including ``Original''; parameters in the second row), averaged over COCO, DIV2K, and Chameleon datasets. Only Affine is seen during training; all other geometric distortions are out-of-distribution.}
  \label{tab:distortions}
  \begin{tabular}{l c c c c c c c}
    \toprule[1.2pt]
    \multirow{2}{*}{Method}
      & Original 
      & Affine
      & Crop 
      & Rotate 
      & Flip 
      & Scale 
      & Elastic \\
      & —
      & $s=10,\ r=60$
      & $[0.3,0.8]$ 
      & $[30^\circ,60^\circ]$ 
      & h / v 
      & $[0.3,1.8]$ 
      & $\alpha=33,\ \sigma=4$ \\
    \midrule
    DCTDWT~\cite{rahman2013dwt}    & 81.25\% & 47.92\% & 49.44\% & 67.19\% & 76.25\% & 81.25\% & 82.29\% \\
    HiDDeN~\cite{zhu2018hidden}    & 96.28\% & 56.47\% & 53.46\% & 56.83\% & 53.42\% & 90.69\% & {\color{BrightBlue}\underline{93.76\%}} \\
    MBRS~\cite{jia2021mbrs}        & {\color{Red}\textbf{100.00\%}} & 49.94\% & {\color{BrightBlue}\underline{95.01\%}} & 58.51\% & 49.89\% & 49.98\% & {\color{Red}\textbf{100.00\%}} \\
    SSL~\cite{fernandez2022watermarkingimagesselfsupervisedlatent} & 84.30\% & 71.86\% & 62.67\% & 74.65\% & 69.60\% & 82.48\% & 74.89\% \\
    RoSteALS~\cite{bui2023rosteals} & {\color{BrightBlue}\underline{99.99\%}} & 50.96\% & 49.99\% & 65.93\% & 48.53\% & {\color{BrightBlue}\underline{99.19\%}} & 53.17\% \\
    TrustMark~\cite{bui2023trustmarkuniversalwatermarkingarbitrary} & 99.98\% & {\color{BrightBlue}\underline{79.28\%}} & 57.14\% & 97.92\% & 63.33\% & 99.09\% & 57.14\% \\
    LaWa~\cite{rezaei2024lawa}     & {\color{Red}\textbf{100.00\%}} & 58.29\% & 62.53\% & {\color{BrightBlue}\underline{98.69\%}} & {\color{BrightBlue}\underline{88.39\%}} & {\color{Red}\textbf{100.00\%}} & {\color{Red}\textbf{100.00\%}} \\
    $\text{MT-Mark}_{\text{(ours)}}$                            & {\color{Red}\textbf{100.00\%}} & {\color{Red}\textbf{98.75\%}} & {\color{Red}\textbf{97.63\%}} & {\color{Red}\textbf{98.99\%}} & {\color{Red}\textbf{99.99\%}} & {\color{Red}\textbf{100.00\%}} & {\color{Red}\textbf{100.00\%}} \\
    \bottomrule[1.2pt]
  \end{tabular}
\end{table*}

\begin{table*}
  \centering
  \small
  \caption{Bit accuracy under signal processing distortions (parameters in the second row), averaged over COCO, DIV2K, and Chameleon datasets. Only Gaussian noise (GN) is seen during training; all other signal-processing distortions are out-of-distribution.}
  \label{tab:pixel_distortions}
  \begin{tabular}{l*{8}{c}c}
    \toprule[1.2pt]
    \multirow{2}{*}{Method}
      & GN      & GB           & BB             & MB         & MeaF         & MedF         & DP             & RE             & RS             \\
      & $\sigma=6$  & $\sigma=6$   & $7\times7$     & $7\times7$ & $7\times7$   & $7\times7$   & $p=0.7$        & $r=0.1$        & $s=0.15$       \\
    \midrule
    DCTDWT~\cite{rahman2013dwt}    & 80.21\% & 81.25\% & 81.25\% & 81.25\% & 81.25\% & 83.33\% & 81.25\% & 69.79\% & 81.25\% \\
    HiDDeN~\cite{zhu2018hidden}    & 83.56\% & 86.73\% & 79.44\% & 76.68\% & 93.33\% & 87.34\% & 78.95\% & 90.36\% & 91.11\% \\
    MBRS~\cite{jia2021mbrs} & 58.06\% & 83.47\% & 64.08\% & 64.14\% & 64.08\% & 67.56\% & {\color{Red}\textbf{99.54\%}} & 99.50\% & 49.98\% \\
    SSL~\cite{fernandez2022watermarkingimagesselfsupervisedlatent} & 62.02\% & 80.64\% & 55.18\% & 70.77\% & 71.63\% & 66.53\% & 50.06\% & 78.64\% & 82.42\% \\
    RoSteALS~\cite{bui2023rosteals} & 99.07\% & 99.15\% & {\color{Red}\textbf{100.00\%}} & 99.14\% & 99.12\% & 99.14\% & {\color{BrightBlue}\underline{98.99\%}} & 98.14\% & {\color{BrightBlue}\underline{96.34\%}} \\
    TrustMark~\cite{bui2023trustmarkuniversalwatermarkingarbitrary} & 85.75\% & 99.93\% & 97.78\% & 99.12\% & {\color{BrightBlue}\underline{99.31\%}} & 97.82\% & 78.89\% & 81.93\% & {\color{Red}\textbf{100.00\%}} \\
    LaWa~\cite{rezaei2024lawa}     & {\color{BrightBlue}\underline{99.57\%}} & {\color{BrightBlue}\underline{99.99\%}} & 90.02\% & {\color{BrightBlue}\underline{99.88\%}} & 99.17\% & {\color{BrightBlue}\underline{99.74\%}} & 96.99\% & {\color{Red}\textbf{100.00\%}} & {\color{Red}\textbf{100.00\%}} \\
    $\text{MT-Mark}_{\text{(ours)}}$      & {\color{Red}\textbf{99.78\%}} & {\color{Red}\textbf{100.00\%}} & {\color{Red}\textbf{100.00\%}} & {\color{Red}\textbf{100.00\%}} & {\color{Red}\textbf{100.00\%}} & {\color{Red}\textbf{100.00\%}} & 97.61\% & {\color{BrightBlue}\underline{99.78\%}} & {\color{Red}\textbf{100.00\%}} \\
    \midrule[1.2pt]
    \multirow{2}{*}{Method}
      & Pos     & Hue          & Sha            & Sat            & Contrast       & Bright         & Color J        & JPEG           & SP             \\
      & $bits=6$& $\pm0.2$     & $\pm0.5$       & $[0.3,1.8]$    & $[0.3,1.8]$    & $[0.3,1.8]$    & $j=0.3$        & $Q=40$         & $p=0.02$       \\
    \midrule
    DCTDWT~\cite{rahman2013dwt}    & 76.04\% & 67.71\% & 81.25\% & 80.21\% & 79.17\% & 81.25\% & 75.00\% & 80.21\% & 77.63\% \\
    HiDDeN~\cite{zhu2018hidden}    & 84.92\% & 62.26\% & 78.61\% & 85.61\% & 83.56\% & 86.67\% & 83.46\% & 85.56\% & 91.62\% \\
    MBRS~\cite{jia2021mbrs} & {\color{BrightBlue}\underline{99.62\%}} & {\color{Red}\textbf{99.38\%}} & {\color{BrightBlue}\underline{99.95\%}} & 99.39\% & 99.48\% & 99.20\% & 99.27\% & {\color{Red}\textbf{100.00\%}} & {\color{BrightBlue}\underline{99.97\%}} \\
    SSL~\cite{fernandez2022watermarkingimagesselfsupervisedlatent} & 81.63\% & 66.17\% & 86.67\% & 90.67\% & 83.37\% & 90.33\% & 80.33\% & 63.33\% & 73.67\% \\
    RoSteALS~\cite{bui2023rosteals} & 94.87\% & 94.58\% & 99.17\% & 99.12\% & 98.83\% & 98.85\% & 98.76\% & 99.03\% & 98.99\% \\
    TrustMark~\cite{bui2023trustmarkuniversalwatermarkingarbitrary} & 98.91\% & 98.53\% & 96.14\% & {\color{BrightBlue}\underline{99.92\%}} & {\color{BrightBlue}\underline{99.72\%}} & {\color{BrightBlue}\underline{99.98\%}} & {\color{BrightBlue}\underline{99.91\%}} & 72.94\% & {\color{Red}\textbf{100.00\%}} \\
    LaWa~\cite{rezaei2024lawa}     & 99.16\% & 96.38\% & {\color{Red}\textbf{100.00\%}} & {\color{Red}\textbf{100.00\%}} & {\color{Red}\textbf{100.00\%}} & {\color{Red}\textbf{100.00\%}} & {\color{Red}\textbf{99.99\%}} & 99.25\% & 96.99\% \\
    $\text{MT-Mark}_{\text{(ours)}}$      & {\color{Red}\textbf{99.83\%}} & {\color{BrightBlue}\underline{99.36\%}} & {\color{Red}\textbf{100.00\%}} & {\color{Red}\textbf{100.00\%}} & {\color{Red}\textbf{100.00\%}} & {\color{Red}\textbf{100.00\%}} & 99.21\% & {\color{BrightBlue}\underline{99.79\%}} & {\color{Red}\textbf{100.00\%}} \\
    \bottomrule[1.2pt]
  \end{tabular}
\end{table*}


\subsection{Robustness and Generalization}

Tables~\cref{tab:distortions,tab:pixel_distortions} report the robustness and generalization performance of our method and representative baselines under a wide range of geometric and signal processing distortions.

Under geometric distortions (\cref{tab:distortions}), traditional transform-based methods such as DCTDWT rely on fixed spatial transforms and therefore struggle to handle large-scale spatial reconfiguration. Deep-learning-based approaches such as HiDDeN and SSL alleviate this limitation to some extent through data-driven learning, but still exhibit noticeable performance degradation under complex transformations such as Crop and Affine. Here, Crop denotes a true cropping operation that changes image size, rather than an occlusion-style pseudo-cropping. MBRS achieves perfect performance under Elastic and Crop distortions; however, it collapses to chance-level accuracy (50\%) under Affine, Flip, and Scale, revealing a strong dependence on specific spatial patterns observed during training. RoSteALS shows limited ability to handle geometric transformations and yields the weakest overall performance in this category. TrustMark and LaWa perform well on selected distortions such as Rotate due to targeted training, but generalize poorly to unseen transformations. For example, LaWa excludes Flip and Affine during training and consequently suffers significant performance drops under these distortions (88.39\% and 58.29\%, respectively), despite near-perfect accuracy on seen cases.

\begin{figure}[!htbp]
  \centering
  \includegraphics[width=\linewidth]{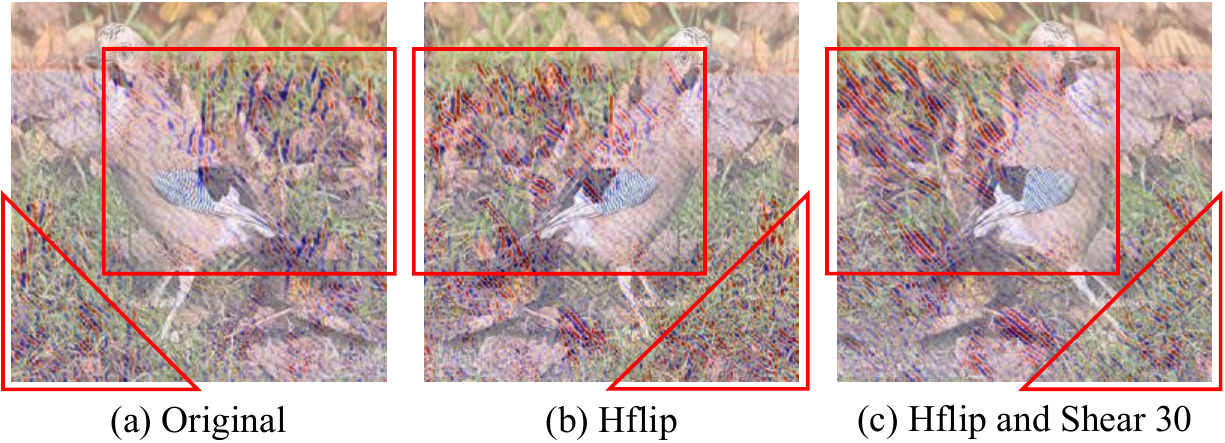} 
  \caption{Qualitative visualization of watermark-focused regions under geometric transformations. (a) Original watermarked image. (b) Watermarked image after horizontal flip (HFlip). (c) Watermarked image after a composite geometric transformation consisting of HFlip followed by an x-axis shear of $30^\circ$. Despite significant spatial reconfiguration, the network consistently focuses on regions carrying watermark information, enabling reliable watermark extraction.}
  \label{att}
\end{figure}

MT-Mark consistently achieves strong performance across all evaluated geometric distortions. Rather than relying on memorizing discrete transformation patterns, MT-Mark learns to maintain stable watermark-related feature responses under continuous spatial transformations. This behavior is further illustrated by the qualitative visualization in \cref{att}. In \cref{att}(a), the network correctly focuses on regions carrying watermark information in the original image. Under a horizontal flip (HFlip), the focused regions remain consistent with the transformed watermark locations, as shown in \cref{att}(b). More importantly, even under a compounded geometric distortion involving HFlip followed by an x-axis shear of $30^\circ$, the network continues to attend to the corresponding watermark-bearing regions in \cref{att}(c). These results indicate that MT-Mark maintains stable watermark-related feature responses under severe spatial transformations, supporting accurate watermark extraction without relying on fixed spatial priors or distortion-specific training.

Under signal processing distortions (\cref{tab:pixel_distortions}), the traditional DCTDWT method improves robustness by adjusting transform-domain coefficients; however, its fixed and hand-crafted design inherently limits the range of distortions it can resist. HiDDeN and SSL adopt distortion-aware training strategies, yet their robustness remains inconsistent, particularly under structured distortions such as filtering and compression. MBRS exhibits strong performance under JPEG compression and color-space perturbations (e.g., Hue, Pos, SP), but is notably brittle under fundamental distortions such as Gaussian noise (58.06\%) and various blur operations. RoSteALS and LaWa perform well under noise and blur conditions, primarily because their training datasets include distortions closely aligned with those used during evaluation. This performance is therefore driven more by training–testing alignment than by genuine generalization capability. TrustMark, despite employing a more sophisticated distortion pipeline, still shows significant degradation in real-world scenarios; for instance, its accuracy drops to 72.94\% under JPEG compression.

 MT-Mark is trained with only Gaussian noise and Affine transformations, yet generalizes effectively to a broad range of unseen signal processing distortions, including blur, filtering, color jitter, and compression. This robustness does not stem from distortion-specific adaptation, but rather from the explicitly collaborative architecture. Through the CIM, the embedder and extractor jointly regulate feature responses, while the AFMM enables dynamic and content-aware feature adjustment. Together, these components stabilize watermark-related features under diverse perturbations, leading to consistently strong extraction performance without reliance on extensive distortion simulation.

\subsection{Ablation Study}

\begin{figure}[t]
  \centering
  \includegraphics[width=\linewidth]{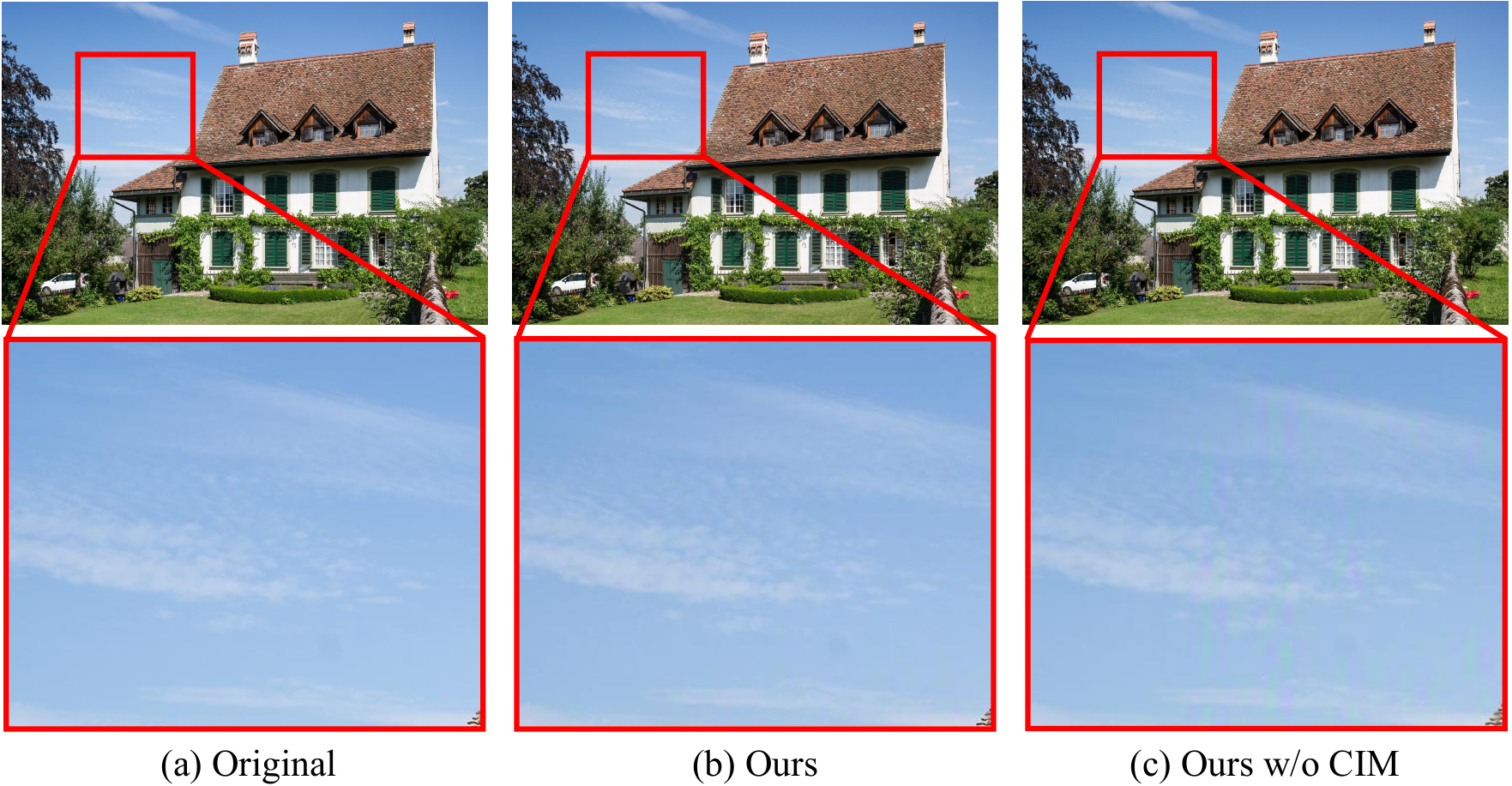}
  \caption{Visual ablation study on the CIM.}
  \label{cim}
\end{figure}

\noindent \textbf{Ablation on  CIM and AFMM.}
We conduct ablation studies to evaluate the contributions of AFMM and CIM. Quantitative results are reported in Table~\cref{gating_interaction_ablation}, where PSNR measures imperceptibility and extraction accuracy is evaluated under combined distortions.
Removing all AFMMs leads to a substantial performance drop, with extraction accuracy decreasing to 64.71\% and PSNR to 18.86,dB, confirming their importance. When only the embedder’s AFMMs are removed, imperceptibility is severely affected, whereas removing the extractor’s AFMMs mainly degrades robustness, indicating their complementary roles.
Disabling CIM while keeping AFMMs also causes a noticeable decline in both PSNR (33.24,dB) and extraction accuracy (91.49\%), suggesting that explicit coordination between embedding and extraction is necessary. This observation is consistent with the qualitative results in \cref{cim}, where removing CIM introduces visible artifacts in the enlarged region in \cref{cim}(c), while the full model remains visually clean in \cref{cim} (b).

\noindent \textbf{Ablation on Distortion and Extractor's multi-scale.} Table~\cref{ablation_distortion_layers_tick} presents ablation results under different distortion training strategies, evaluated on combined distortions.
Training without distortions yields poor robustness (72.19\%). Models trained with a single distortion type exhibit limited generalization, while training with both Affine and Gaussian noise achieves the best performance (99.07\%), comparable to training with all distortions (98.91\%). This confirms that robust performance can be achieved without exhaustive distortion augmentation. 
Additionally, removing the multi-scale design from the extractor leads to a 6.7\% drop in average extraction accuracy across distortions, demonstrating its necessity for robustness.

\begin{table}
\centering
\small
\caption{Ablation Results on AFMM and CIM Modules Reflecting Imperceptibility and Robustness.}
\label{gating_interaction_ablation}
\begin{tabular}{lccccc}
\toprule[1.2pt]
                & \textbf{(a)} & \textbf{(b)} & \textbf{(c)} & \textbf{(d)} & \textbf{(e)} \\
\midrule
\textbf{Emb.'s AFMMs} & $\checkmark$ & $\times$      & $\checkmark$  & $\times$      & $\checkmark$ \\
\textbf{Ext.'s AFMMs} & $\checkmark$ & $\checkmark$  & $\times$      & $\times$      & $\checkmark$ \\
\textbf{CIM}          & $\checkmark$ & $\checkmark$  & $\checkmark$  & $\checkmark$  & $\times$     \\
\midrule
\textbf{PSNR}         & \textcolor{red}{\textbf{38.96}} & 21.09 & 26.44 & 18.86 & \textcolor{blue}{\textbf{33.24}} \\
\textbf{Combine (\%)}& \textcolor{red}{\textbf{99.36}} & 87.24 & 76.58 & 64.71 & \textcolor{blue}{\textbf{91.49}} \\
\bottomrule[1.2pt]
\end{tabular}
\end{table}


\begin{table}[t]
\centering
\small
\caption{Ablation on distortion training strategies. Models were trained with different strategies and evaluated under combined distortions.}
\label{ablation_distortion_layers_tick}
\begin{tabular}{lccccc}
\toprule[1.2pt]
                & \textbf{(a)} & \textbf{(b)} & \textbf{(c)} & \textbf{(d)} & \textbf{(e)} \\
\midrule
\textbf{Affine}   & $\checkmark$ & $\times$ & $\checkmark$ & $\times$ & $\checkmark$ \\
\textbf{Gaussian} & $\checkmark$ & $\times$ & $\times$      & $\checkmark$ & $\checkmark$ \\
\textbf{All distortions} & $\times$ & $\times$ & $\times$ & $\times$ & $\checkmark$ \\
\midrule
\textbf{Combine (\%)}  & \textcolor{red}{\textbf{99.07}} & 72.19 & 83.12 & 81.65 & \textcolor{blue}{\textbf{98.91}} \\
\bottomrule[1.2pt]
\end{tabular}
\end{table}

%% file: sec/5_con.tex



\section{Conclusion}

We present MT-Mark, an architecture-level redesign for robust image watermarking that establishes explicit collaboration between watermark embedding and extraction. 
At its core, a collaborative interaction mechanism coordinates the two components in a mutual-teacher manner, while adaptive feature modulation serves as the execution interface that enables fine-grained, content-aware adjustment under this coordination.
This design allows robustness and imperceptibility to emerge from coordinated representation learning rather than exhaustive distortion enumeration. 
Extensive experiments demonstrate that MT-Mark consistently outperforms state-of-the-art methods under geometric and signal-processing distortions, and remains highly robust under semantic editing operations and image-to-video generation, despite being trained with only a minimal set of representative distortions. 
Ablation studies further validate that both the collaborative mechanism and its guided feature modulation are essential for achieving strong generalization and high perceptual fidelity.